\pdfoutput=1

\documentclass[11pt]{article}

\usepackage[final]{acl}

\usepackage{times}
\usepackage{latexsym}

\usepackage[T1]{fontenc}

\usepackage[utf8]{inputenc}

\usepackage{microtype}

\usepackage{inconsolata}

\usepackage{graphicx}
\usepackage{booktabs}
\usepackage{multirow}
\usepackage{makecell}
\usepackage{array}
\usepackage{xcolor}
\usepackage{fontawesome5}
\usepackage{hyperref}
\usepackage{float}
\usepackage{afterpage}
\usepackage{pgfplots}
\usepackage{pgfplotstable}
\usepackage{xcolor}
\usepackage{enumitem}
\pgfplotsset{compat=1.17}

\usepackage{amsmath}
\usepackage{amssymb}
\usepackage{mathtools}
\usepackage{amsthm}
\usepackage{tcolorbox}
\usepackage{bbm}
\usepackage{fontawesome5}
\usepackage{multicol}
\usepackage{amsmath}
\usepackage{subcaption}
\tcbuselibrary{skins}
\newcommand{\hdr}[3][0ex]{%
  \makecell[tc]{%
    \\[-0.25ex]\raisebox{#1}[0pt][0pt]{#2}\\[-0.3ex]%
    \raisebox{-#1}[0pt][0pt]{#3}}}

\usepackage{amsmath,amsfonts,bm}

\definecolor{tablecolor}{rgb}{0.8,0.8,0.8}

\newcommand{\vparam}{\vtheta}

\newcommand{\vocab}{\mathcal{V}}

\newcommand\cut[1]{}

\newcommand{\squishlist}{
   \begin{list}{$\bullet$}
    { \setlength{\itemsep}{0pt}      \setlength{\parsep}{3pt}
      \setlength{\topsep}{3pt}       \setlength{\partopsep}{0pt}
      \setlength{\leftmargin}{1.5em} \setlength{\labelwidth}{1em}
      \setlength{\labelsep}{0.5em} } }

\newcommand{\squishlisttwo}{
   \begin{list}{$\bullet$}
    { \setlength{\itemsep}{0pt}    \setlength{\parsep}{0pt}
      \setlength{\topsep}{0pt}     \setlength{\partopsep}{0pt}
      \setlength{\leftmargin}{2em} \setlength{\labelwidth}{1.5em}
      \setlength{\labelsep}{0.5em} } }

\newcommand{\squishend}{
    \end{list}  }

{}
{}
{}

\newcommand{\real}{\mbox{$\mathbb{R}$}}

\newcommand{\myvec}[1]{\mathbf{#1}}

\newcommand{\vtheta}{\mathbf{\boldsymbol{\theta}}}

\newcommand{\va}{\mbox{$\myvec{a}$}}

\newcommand{\vs}{\mbox{$\myvec{s}$}}

\newcommand{\vu}{\mbox{$\myvec{u}$}}

\newcommand{\vP}{\mbox{$\myvec{P}$}}

\usepackage[capitalize,noabbrev]{cleveref}

\theoremstyle{plain}

\theoremstyle{definition}

\theoremstyle{remark}

\usepackage[textsize=tiny]{todonotes}

\newcommand*\iftodonotes{\if@todonotes@disabled\expandafter\@secondoftwo\else\expandafter\@firstoftwo\fi}  %
\makeatother

\title{From Problem-Solving to Teaching Problem-Solving: \\
Aligning LLMs with Pedagogy using 
Reinforcement Learning
}

\author{
    David Dinucu-Jianu$^{\ast 1}$ \quad
    Jakub Macina$^{\ast 1, 2}$ \quad 
    Nico Daheim$^{1, 3}$ \quad \\
    \textbf{
    Ido Hakimi$^{1, 2}$ \quad
    Iryna Gurevych$^3$ \quad
    Mrinmaya Sachan$^{1}$
    } \\ \text{} \\
  $^{1}$Department of Computer Science, ETH Zurich \quad
  $^2$ETH AI Center \\
  $^{3}$Ubiquitous Knowledge Processing Lab (UKP Lab), Department of Computer Science, \\ Technical University of Darmstadt and National Research Center \\ for Applied Cybersecurity ATHENE, Germany 
}
  
\begin{document}
\maketitle
\def\thefootnote{*}\footnotetext{Equal contribution.}\def\thefootnote{\arabic{footnote}}

\begin{abstract}
Large language models (LLMs) can transform education, but their optimization for direct question-answering often undermines effective pedagogy which requires strategically withholding answers. To mitigate this, we propose an online reinforcement learning (RL)-based alignment framework that can quickly adapt LLMs into effective tutors using simulated student-tutor interactions by emphasizing pedagogical quality and guided problem-solving over simply giving away answers. We use our method to train a 7B parameter tutor model without human annotations which reaches similar performance to larger proprietary models like LearnLM. We introduce a controllable reward weighting to balance pedagogical support and student solving accuracy, allowing us to trace the Pareto frontier between these two objectives. Our models better preserve reasoning capabilities than single-turn SFT baselines and can optionally enhance interpretability through thinking tags that expose the model's instructional planning.

\end{abstract}

\hspace{0em}\includegraphics[width=0.9em,height=0.9em]{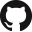}\hspace{.25em}\parbox{\dimexpr\linewidth-2\fboxsep-2\fboxrule}{\href{https://github.com/eth-lre/PedagogicalRL}{https://github.com/eth-lre/PedagogicalRL}}

\begin{figure}[t]
  \centering
  \includegraphics[width=0.48\textwidth]{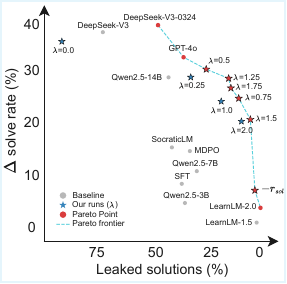}
  \caption{LLM tutoring forms a multi-objective scenario in which LLM tutors should increase the student's solve rate (y-axis) while minimizing solution leakage (x-axis). Here, the $\Delta$ solve rate measures how often a student can solve a problem before and after the dialog with a tutor and leaked solutions measures how often the tutor tells the solution to the student. Our RL-trained Qwen-2.5-7B models with varying penalty $\lambda$ are on the Pareto-front and match the performance of specialized closed-source models when tutoring on Big-Math. 
  }
  \label{fig:pareto_frontier}
\end{figure}

\section{Introduction}
\label{introduction}
Large Language Models (LLMs) hold significant promise in education, particularly as personalized tutors capable of guiding students individually through problems. Recent advances have demonstrated remarkable LLM performance in math and science~\cite{alphageometry2, medgemini}. However, deploying LLMs effectively as educational tutors involves more than excelling on benchmarks~\cite{tack2022ai, beyond_final_answers}. To be truly effective, a tutor must facilitate learning by guiding students toward independently constructing correct solutions rather than simply revealing the answers. We refer to this shift from assistant to tutor as pedagogical alignment.

Achieving robust pedagogical alignment remains an open challenge~\cite{MathTutorBench,maurya-etal-2025-unifying}. Approaches that rely on supervised fine-tuning (SFT)~\cite{stepwise,kwon-etal-2024-biped} can suffer from generalization issues
while existing RL-based techniques typically depend on costly, and often proprietary, preference annotations~\cite{learnlm} or require a much larger model as a source of training data of tutor responses~\cite{naive_rl_approach,scarlatos2025dpomathdial}. Due to these limitations, these prior works have largely focused on single-turn feedback, which fails to capture the multi-turn dynamics that are essential for effective tutoring.

\begin{figure*}[t!]
  \centering
  \includegraphics[width=0.87\textwidth]{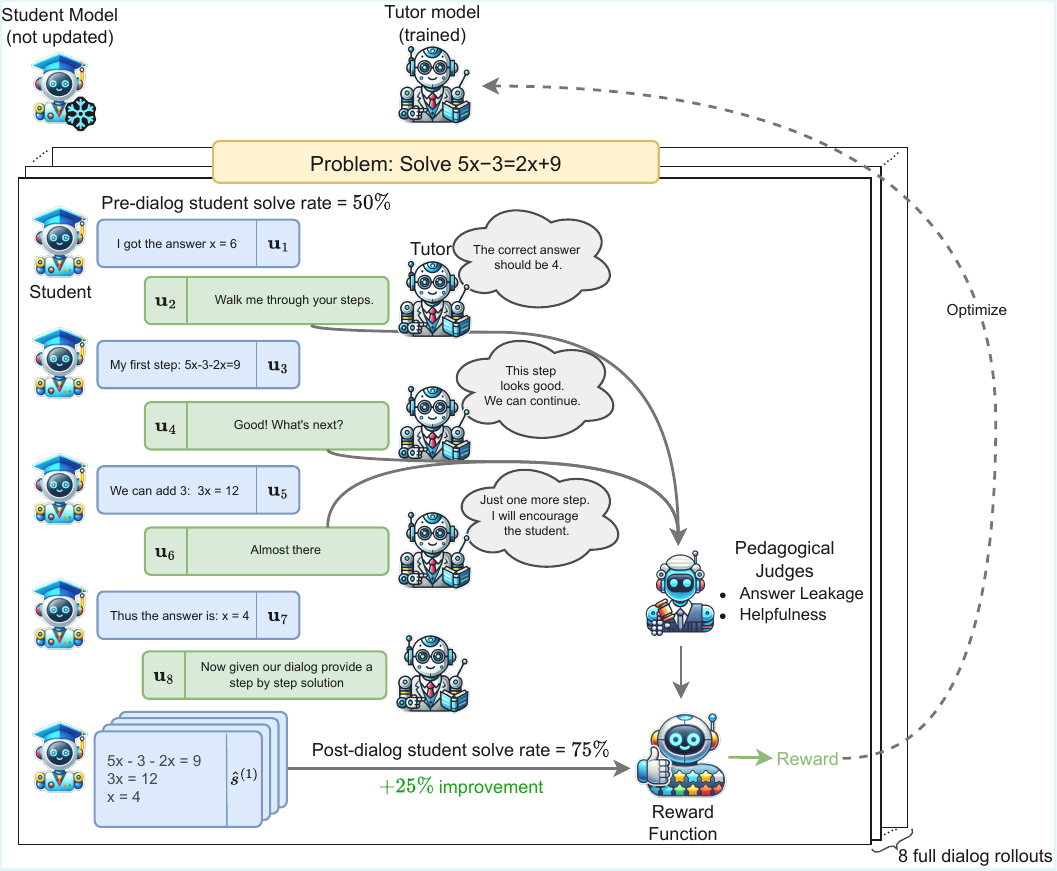}
  \caption{Workflow of our RL framework. First, we perform multiple complete student-tutor conversation simulations. After each conversation ends, the reward is computed: 1) post-dialog student solve rate (success) conditioned on the dialog, and 2) the pedagogical quality of the tutor guidance throughout the conversation. This setup uses data from the current tutor model (is on-policy) and does not use offline static dialog data (is online).
  }\label{fig1:framework_diagram}
\end{figure*}

To address these gaps, we propose a multi-turn reinforcement learning (RL) method that enables the model to learn directly from its own dialogs with a student to find optimal teaching strategies. 
Grounded in mastery learning and active teaching principles~\cite{chi2014icap, freeman2014active}, our system simulates multi-turn interactions on challenging problems from Big-Math~\cite{bigmathrl}, with the tutor LLM using Socratic questioning~\cite{shridhar-etal-2022-automatic} and targeted hints instead of handing out solutions. We design reward functions that mirror 
authentic long-term learning outcomes, namely, how often a student can solve a problem after a dialog with the tutor and how much the tutor follows sound pedagogical principles throughout the full conversation.
Our key contributions are the following:
\begin{itemize} 

\item \textbf{Cost-efficient training via synthetic student–tutor interactions:} Our online RL method replaces the need for expensive human-annotated data with a synthetic data pipeline, enabling a 7B Tutor Model to almost match the performance of LearnLM.

\item \textbf{Controllable pedagogy–accuracy trade-off:} Our method enables explicit control over the balance between pedagogical support and student answer correctness by adjusting a penalty weight to navigate a Pareto frontier.

\item \textbf{Preservation of reasoning capabilities:} Our approach maintains performance across standard reasoning benchmarks, unlike prior methods such as SocraticLM~\cite{socraticlm}.
Evaluations on MMLU, GSM8K, and MATH demonstrate that pedagogical alignment does not come at the cost of reasoning ability.

\end{itemize}

\section{Related Work}

\subsection{\bf LLMs for Dialog Tutoring}
While effective human tutors not only provide answers but more importantly scaffold the learning of students, LLMs are predominantly trained for providing answers which limits their tutoring capabilities~\cite{tack2022ai,macina-etal-2023-opportunities}. 
Hence, various approaches have been proposed to improve their pedagogical skills.

Arguably the simplest is prompt engineering, where pedagogical criteria are encoded in the prompt, for example, for asking questions~\citep{sonkar-etal-2023-class,puech2025pedagogicalsteeringlargelanguage} or detecting mistakes~\citep{wang-etal-2024-bridging} but it is tedious and sensitive to changes~\citep{jurenka2024towards}.

A more robust alternative is to use gradient-based updating, for example, SFT on teacher-student dialogs.
However, this is challenging because only a few high-quality tutoring datasets exist publicly, for example, MathDial which is semi-synthetically created by pairing LLM students with real teachers for solving math problems~\citep{macina2023mathdialdialoguetutoringdataset}.
Hence, many works resort to synthetic data~\citep{wang-etal-2024-book2dial}. For example, SocraticLM~\citep{socraticlm} is trained on 35k math tutoring dialogs created using a multi-agent setting and TutorChat~\citep{tutorchat} is trained using 80k synthetic teacher–student conversations grounded in textbooks.
Larger scale approaches in industry, such as, LearnLM~\cite{jurenka2024towards} use a mixture of synthetic and human-collected data but this requires substantial resources.

Finally, recent works use Reinforcement Learning from Human Feedback (RLHF)~\citep{ouyang2022training}, for example, to improve next tutor dialog act prediction~\citep{naive_rl_approach} or to improve math tutors by turn-level rewards using GPT-4-generated preference data~\citep{scarlatos2025dpomathdial}.
However, it is unclear how single-turn synthetic data translates to tutoring more complex multi-turn conversations.

Prior works treat tutoring as an offline off-policy problem by relying on large-scale synthetic or proprietary data which introduces exposure bias~\citep{ross2010efficient, Ranzato2016} as the tutor does not learn from its own interactions during training.
In contrast, our work adopts an online on-policy setup where the model is trained on its own interactions throughout training.

\subsection{\bf Dialog as RL Task \& Verifiable Rewards}
Previous work has commonly framed educational dialog as a next teacher utterance generation task, where the teacher's last turn serves as a ground truth response and the dialog history serves as context~\cite{macina2023mathdialdialoguetutoringdataset}. 
However, a dialog is inherently a multi-turn interaction towards a goal (e.g. student learns to solve a problem) and single-turn methods limit the model's ability to plan across multiple turns to achieve longer-term goals. Effective tutoring, however, is a sequential, adaptive and goal-directed process with the aim of helping a student not only solve a current problem, but also learn to solve similar problems.
To address this problem, formulating
dialog as an RL problem might be helpful which has been explored outside of tutoring recently~\cite{li2017learning,shani2024mtpo,xiong2025building,li2025multi-turn-survey}. 

In general, RL learns optimal actions by collecting a numerical reward from the environment which provides a natural framework for aligning LLM behavior with pedagogical goals by assigning rewards to complete conversations rather than to isolated turns. In LLMs, RL has been successfully used to align with human feedback~\cite{ouyang2022training} and to improve reasoning via verifiable rewards~\cite{deepseekmath,lambert2024t,wang2025ragenunderstandingselfevolutionllm}. 

Standard on-policy algorithms like Proximal Policy Optimization (PPO)~\cite{schulman2017proximal} have been crucial for the success of human-preference alignment in GPT models. Direct Preference Optimization (DPO)~\cite{dpo} has emerged as a simpler alternative without the requirement of a reward model that allows fine-tuning on offline pairwise preference data. Extensions of DPO to multi-turn settings, such as multi-turn DPO (MDPO), commonly mask user turns to optimize only over assistant responses~\cite{xiong2025building}. Recent algorithms such as MTPO~\cite{shani2024mtpo} and REFUEL~\cite{gao2025regressing} compare pairs of entire conversations rollouts to improve over DPO.
Access to verifiable rewards has been crucial for scaling RL training for LLMs, for example, by comparing to a reference solution~\cite{deepseekmath,deepseek_r1} or executing programs~\cite{lambert2024t}. While these methods have been used to improve reasoning, pedagogical criteria have largely been neglected.

Our work builds upon a line of research formulating a dialog as an RL problem in a synthetic tutor-student environment. By integrating verifiable correctness rewards with pedagogical rubrics, we explore the control of the trade-off between instruction support and answer accuracy.

\section{Pedagogical Principles}\label{sec:ped-principles}
Effective teaching is not only about providing answers but rather about fostering student learning through scaffolding guidance. 
Here, scaffolding means actively engaging students in problem solving~\cite{chi2014icap,freeman2014active} using questions, hints, and nudges. 

\textbf{Avoiding Answer Leakage:} A key element is to actively engage students in problem solving instead of letting them passively consume correct answer, which does not lead to learning.
Therefore, we discourage the tutor from presenting complete solutions.
Instead, they should guide students through Socratic questioning, hints, or targeted feedback. This mirrors constraints from prior related work, such as the role of a dean persona~\cite{socraticlm}.

\textbf{Helpfulness:}
The tutor should guide the student with constructive and contextual appropriate support in the right teacher tone. The tutor violates this principle if they provide full answers or dominate the conversation and it is similar to targetedness in prior work~\cite{stepwise}. Moreover, tutors should be responsive and encouraging, reflecting the tone of real teachers~\cite{tack2022ai}.

\section{Dialog Tutoring as Multi-Turn RL} 
We consider multi-turn conversations $(\vu_1, \dots, \vu_T)$ made up of a sequence of utterances $\vu_t \in \vocab^\ast$ taken by either the student or a teacher, both simulated by an LLM.
In our training runs, it is decided by random choice who starts the conversation, as detailed in~\cref{sec:details}.
The goal of the student is to solve a problem $\vP \in \vocab^\ast$ which has a unique known numerical solution $s \in \real$. The objective of the LLM tutor is to guide the student toward the solution $s$ by generating a new $\vu_t$ given the context $\vu_{<t}$.
The conversation ends when the tutor considers it finished or after a fixed number of turns.
We use autoregressive LLM-based tutors, parameterized by neural network weights $\vparam$, to generate outputs by sampling from the model distribution \begin{align*}
    p_{{\vparam}}(\vu_t\mid \vu_{<t}) = \prod_{n=1}^{|\vu_t|}p_{{\vparam}}([u_t]_n \mid [\vu_t]_{<n}, \vu_{<t}),
\end{align*}
where $[u_t]_n$ is the $n$-th token of the output sequence $\vu_t$. In~\cref{sec:ped-principles} we define the pedagogical principles that the generated utterances should fulfill.

Learning ${\vparam}$ can then also be framed as an RL problem under the lens of Markov Decision Processes (MDP) for which we re-define the previously introduced quantities in common notation.
To be precise, for a given position $t$ in the dialog, we define the state to be $\vs_t \coloneqq \vu_{<t}$ and the action to be $\va_t \coloneqq \vu_t$, i.e. the current state in the conversation is fully captured by the sequence of previous utterances and the action is the next utterance.
The transition dynamics are defined by sequentially appending each new utterance (or action) $\va_t$ to the existing conversation history (or state $\vs_t$) to form the new state $\vs_{t+1}$. If $\va_t$ is a tutor utterance, it is sampled from the tutor's policy; if it is a student utterance, it is sampled from a fixed student LLM conditioned on $\vs_t$. 
Since the student model is stochastic, the transition dynamics are non-determinsitic from the tutor’s perspective, different from standard RLHF which often assumes deterministic environments.

Then, the goal is to learn the tutor policy $\pi_{\vparam} \coloneqq p_{\vparam}$ such that sampled responses \begin{equation}
    \va_t \sim \pi_{\vparam}(\cdot\mid \vs_t)
\end{equation} fulfill the desiderata in ~\cref{sec:ped-principles}.
We achieve this by defining rewards $r(\va_T, \vs_T)$ that are assigned at the end of a conversation to full sequences $\va_T$ based on the context $\vs_T$.
That is, we define rewards at the level of the full conversation rather than assigning them to individual 
turns.
Furthermore, we also sample $\va_t$ directly from the current policy $\pi_{\vparam}$ at the given training iteration. The on-policy approach means we update the current policy $\pi_{\vparam}$ and subsequent dialogs are generated from the newly updated model. This is different from DPO-based approaches, which use static data. There, the model is always conditioned on context from an older checkpoint. 
Instead, we use online RL and avoid such context drift by conditioning on context generated with the current model checkpoint.

\subsection{Rewarding LLM Tutor Pedagogy}
Our reward design follows the pedagogical principles laid out in~\cref{sec:ped-principles}.
This means that we aim to fulfill two goals: the student should be able to successfully solve $\vP$ after the dialog and the actions $\va_t$ generated using the policy $\pi_{\vparam}$ should have high pedagogical quality and, for example, not just solve the problem for a student. 

We judge solution correctness by sampling multiple final answers $\widehat{s}^{(1)}, \widehat{s}^{(2)}, \dots, \widehat{s}^{(K)}$ from the student model conditioned on $\va_T$ and $\vs_T$ 
and compute an empirical expected correctness across these solutions called post-dialog solve rate: \begin{equation}
    r_{\text{sol}}(\va_T\mid \vs_T) = \frac{1}{K}\sum_{k=1}^K \mathbbm{1}[\widehat{s}^{(k)} = s],
\end{equation}
where $s$ is the ground-truth solution, as a verifiable outcome reward~\citep{deepseek_r1}.

We judge pedagogical quality (defined in ~\cref{sec:ped-principles}) using LLM judges $J_1, J_2, \dots, J_M$ to prevent overfitting on one specific judge model~\cite{judge_ensambles2024}.
We prompt the judge models independently to evaluate the full conversation and then only consider a conversation accepted if all judges accept it by measuring: \begin{equation}
    r_{\text{ped}}(\va_T\mid\vs_T) = \prod_{m=1}^M\mathbbm{1}[J_m(\va_T, \vs_T) = \text{accept}].
\end{equation}
Altogether, we combine these rewards as: \begin{align}
    r(\va_T\mid \vs_T) &= r_{\text{sol}}(\va_T\mid \vs_T) \nonumber
    \\
    &\quad + (r_{\text{ped}}(\va_T\mid \vs_T) - 1) \cdot \lambda
\end{align}
given a penalty $\lambda \geq 0$ which is a hyperparameter.
The penalty gets subtracted only if any of the pedagogical judges ($r_{ped}=0$) do not accept the conversation.

Intuitively, this provides a way of trading off solution correctness indicated by $r_{\text{sol}}$ against pedagogy measured by $r_{\text{ped}}$. 
If we only care about solution correctness, we can choose $\lambda = 0$ but would expect low pedagogy and many answers given away by the tutor.
On the other hand, if we send $\lambda \rightarrow \infty$, only pedagogy matters which might mean that the student solves fewer problems but actually learns how to solve them.
In between, various trade-offs can be explored.
Finally, we also try a version called \textit{hard} -- if the conversation is not accepted by at least one judge ($r_{\text{ped}}=0$), the overall reward is set to a fixed penalty $-\lambda$ to reflect pedagogical acceptance as a hard prerequisite.

\begin{table*}[h]
\centering
\begin{tabular}{l r r | l r r}
\toprule
\multicolumn{3}{c|}{\textbf{Training Set}} & \multicolumn{3}{c}{\textbf{Test Set}}\\
\midrule
\textbf{Dataset} & \textbf{Samples} & \textbf{Solve Rate (\%)} & \textbf{Dataset} & \textbf{Samples} & \textbf{Solve Rate (\%)} \\
\midrule
\texttt{Big\_math}   & 3360 & 23.56 & \texttt{Big\_math}   & 177 & 24.86 \\
\texttt{Cn\_k12}     & 3324 & 22.11 & \texttt{Cn\_k12}     & 168 & 22.34 \\
\texttt{Math}        & 1264 & 27.40 & \texttt{Math}        & 57  & 23.93 \\
\texttt{Aops\_forum} & 1263 & 10.13 & \texttt{Aops\_forum} & 56  & 10.07 \\
\texttt{Omnimath}    & 374  & 12.57 & \texttt{Omnimath}    & 22  & 15.41 \\
\texttt{Openmath}    & 315  & 38.18 & \texttt{Openmath}    & 13  & 36.30 \\
\texttt{Gsm8k}       & 100  & 36.30 & \texttt{Gsm8k}       & 7   & 32.14 \\
\midrule
\textbf{Total} & \textbf{10 000} & & \textbf{Total} & \textbf{500} & \\
\bottomrule
\end{tabular}
\caption{Composition of training and test datasets with the student model solve rates (pre-dialog).}\label{tab:train_test_composition}
\end{table*}

\begin{figure}[h]
    \centering
    \hspace{-2em}
    \includegraphics[width=0.45\textwidth]{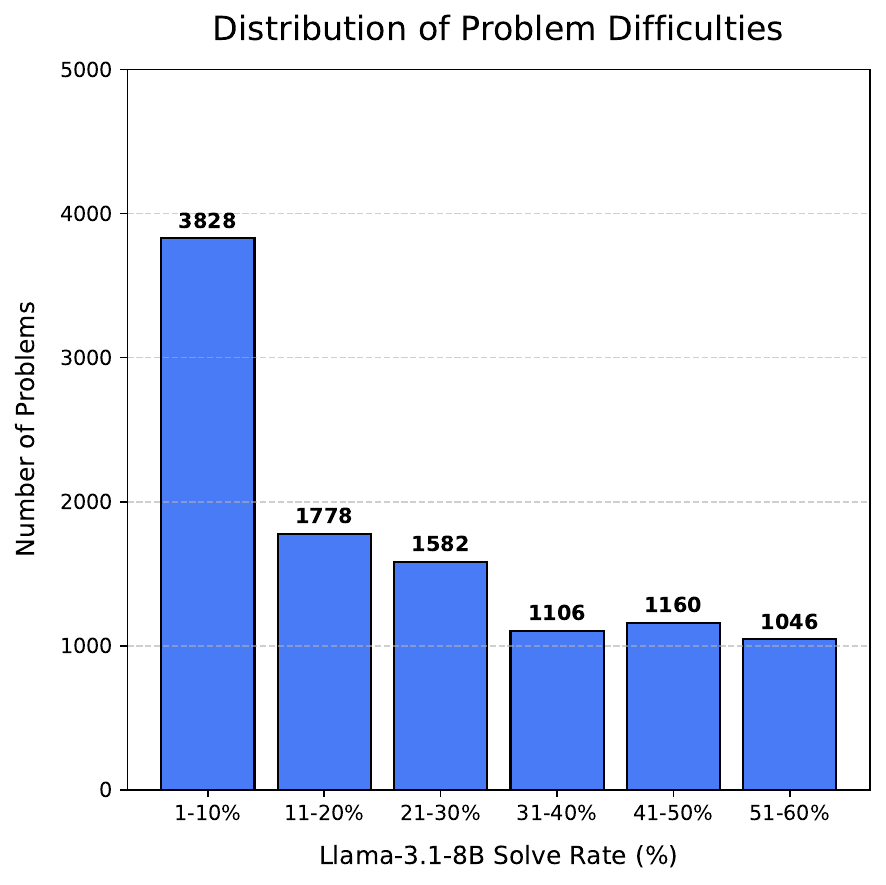}
    \caption{Distribution of problem difficulties in our dataset (solve‑rate buckets obtained with our student model \texttt{Llama‑3.1‑8B‑Instruct}). 
    The dataset contains mostly hard (1-10\% solve rate) problems. This ensures each item requires meaningful guidance from the tutor model rather than being trivial for our student model.
    }
    \label{fig:difficulty_distribution}
\end{figure}

\section{Experiments}\label{experiments}

\subsection{Details on the RL Environment}
\label{sec:details}
Our simulated environment is designed to mimic multi-turn interactions between a student and a tutor. Each episode is seeded with the problem $\vP$ that the student is trying to solve. 
An overview of the environment and an example of a conversation are in~\cref{fig1:framework_diagram}. 
The environment supports two types of common educational interactions which differ in who starts the conversation.
One option is to let the LLM student provide an attempted solution which may be correct, incorrect, or partially correct. Then, the tutor continues the conversation based on the initial attempted solution.
Another scenario is that the tutor initiates the dialog and elicits a solution from the student LLM.
We uniformly sample from the two scenarios in our experiments.

Furthermore, to enable the tutor model to plan and generate more targeted responses, we adopt thinking tags~\cite{openai_o1, deepseek_r1} where the tutor can plan the response.
This content is hidden to the student LLM.

\subsection{Dataset}
We evaluate our framework on BigMath~\cite{bigmathrl} which contains multi-step math problems. The dataset is annotated with the solve rate of \texttt{Llama-3.1-8B-Instruct} with chain-of-thought prompting~\cite{cot}. We only use problems with a single numerical answer and medium-to-high difficulty, i.e., a solve rate by student model between 1\% and 60\% out of $64$ samples. Dataset details are in Table~\ref{tab:train_test_composition} and a distribution over problem difficulties is in \cref{fig:difficulty_distribution}.
We partition this dataset into 10,000 training samples and 500 test samples. On the train dataset, our student model \texttt{Llama-3.1-8B-Instruct} achieves an average pre-dialog solve rate of $25\%$ while 
\texttt{Qwen2.5-7B-Instruct} achieves $66\%$.

To evaluate our models, we adopt following test beds:

\paragraph{Held‑out BigMath (in‑domain):}  
          We first report results on the 500 held‑out BigMath problems. This mirrors the training setting and verifies whether our RL pipeline optimizes the intended conversational rewards. 
          Our main metrics are the \textit{$\Delta$ Solve rate (\%)} and \textit{Leaked Solution (\%)}. \textit{$\Delta$ Solve rate (\%)} measures improvement in the student's problem-solving success after dialog. It is the difference between pre-dialog solve rate measured using chain-of-thought accuracy and the post-dialog solve rate, with both computed in comparison to the ground truth solution $s$. \textit{Leaked Solution (\%)} is a portion of conversations where the tutor gives away the solution to the student assessed by an LLM judge (prompt in Figure~\ref{fig:leakage-prompt}).
          
\paragraph{MathTutorBench (out‑of‑domain):}  
      We additionally evaluate on the independent MathTutorBench benchmark \cite{MathTutorBench}, which provides several automatic metrics for tutor quality. We mainly focus on those metrics that rely on the benchmark’s learned \emph{Pedagogical Reward Model (Ped‑RM)}, as they directly reflect the quality of scaffolding and other pedagogical best practices. Note that the \textit{Ped-RM} score is only used for evaluation across this paper and not as part of the reward.
      
\paragraph{Reasoning Benchmarks:} 
    Finally, to ensure that tutor specialization does not degrade reasoning ability, we also report performance on the general-purpose benchmarks \textsc{MMLU} \cite{mmlu}, \textsc{GSM8K} \cite{gsm8k}, and \textsc{MATH500} \cite{verifystepbystep}.

\subsection{Implementation Details}
We use Group Relative Policy Optimization (GRPO)~\cite{deepseekmath} for model optimization.
For each problem, we simulate 8 complete student–tutor dialogs (rollouts). A single reward score reflecting student success and pedagogical quality of the entire dialog is assigned at the end of each simulation.
We follow the standard GRPO to normalize each dialog reward within each group to obtain dialog-level advantages. The advantages are computed by comparing the reward of a sampled dialog with others in its group. Then dialog-level advantages are propagated to the token-level by adjusting the likelihood of generating each token. We mask the student turns to only optimize over tutor responses. We treat all tutor utterances equally and apply no discounting factor.
The maximum number of total turns is set to 16. Moreover, we use a reward for template following based on the success of~\citet{deepseek_r1}, see details in Appendix~\ref{app:template-reward}.
To compute $r_{\text{ped}}$, we use two judge prompts: Answer Leakage in Figure~\ref{fig:leakage-prompt} and Helpfulness in Figure~\ref{fig:constructive-style-prompt}, and sample twice from each.

\begin{table*}[!htb]
    \centering
    \small
    \begin{tabular}{lccc}
    \toprule
    Model & $\Delta$ Solve rate (\%) $\uparrow$ & Leak Solution (\%) $\downarrow$ & Ped-RM micro/macro $\uparrow$ \\
    \midrule
    \multicolumn{4}{l}{\textit{Our Models}} \\
    \cmidrule(lr){1-4}

    Qwen2.5-7B-RL--$\lambda$=0.0 & 36.2 & 89.5 & -2.8/-3.2 \\
    Qwen2.5-7B-RL--$\lambda$=0.25 & 29.3 & 32.0 & 2.3/1.8 \\
    Qwen2.5-7B-RL--$\lambda$=0.5 & 30.9 & 25.1 & 2.7/1.5 \\
    Qwen2.5-7B-RL--$\lambda$=0.75 & 25.3 & 10.6 & 3.9/3.2 \\
    Qwen2.5-7B-RL--$\lambda$=1.0 & 24.7 & 18.4 & 3.2/2.2 \\
    Qwen2.5-7B-RL--$\lambda$=1.25 & 29.1 & 15.1 & 3.6/3.1 \\
    Qwen2.5-7B-RL--$\lambda$=1.5 & 21.2 & 5.4 & 4.4/4.0 \\
    \;\; + think & 17.0 & 7.4 & 4.9/4.6 \\
    Qwen2.5-7B-RL-hard--$\lambda$=1.0 & 12.6 & 5.3 & 4.2/3.4 \\
    \;\; + think & 20.5 & 6.9 & 4.3/4.9 \\
    \;\; -- $r_{\text{sol}}$ & 7.6 & 3.4 & 3.9/3.1 \\

    \midrule
    \multicolumn{4}{l}{\textit{Baselines – Specialized Tutoring Models}} \\
    \cmidrule(lr){1-4}
    SocraticLM & 15.9 & 40.4 & 1.7/1.7 \\
    Qwen2.5-7B-SFT & 8.9 & 36.0 & -0.3/-0.7 \\
    Qwen2.5-7B-MDPO & 16.4 & 35.6 & 0.2/-0.3 \\
    LearnLM 1.5 Pro Experimental & 1.5 & 2.6 & 5.9/5.3 \\
    LearnLM 2.0 Flash Experimental & 4.3 & 0.9 & 6.8/6.4 \\
    \midrule
    \multicolumn{4}{l}{\textit{Open-Weights Models}} \\
    \cmidrule(lr){1-4}
    Qwen2.5-3B-Instruct & 5.2 & 34.6 & -1.6/-1.7 \\
    Qwen2.5-7B-Instruct & 11.3 & 29.3 & -0.2/-0.5 \\
    Qwen2.5-14B-Instruct & 29.3 & 41.9 & -0.6/-1.2 \\
    Qwen2.5-72B-Instruct & 38.7 & 61.0 & 1.8/-0.4 \\
    DeepSeek V3-0324 & 39.3 & 46.6 & -1.5/-0.8 \\
    \midrule
    \multicolumn{4}{l}{\textit{Closed-Source Models}} \\
    \cmidrule(lr){1-4}
    GPT-4o-2024-11-20 & 33.1 & 35.2 & 1.5/-0.3 \\
    \bottomrule
    \end{tabular}
    \caption{Main results based on in-domain test set. $\Delta$ Solve rate refers to the difference between pre- and post-dialog student solve rate. An independent model (\texttt{Gemma3-27B}) judges the leakage solution. The Per-RM score is only used for evaluation. Macro refers to averaging per conversation while micro uses averaging of all individual scores. 
    }
    \label{tab:result_vs_multi}
\end{table*}

\subsection{Models}

We use \texttt{Qwen2.5-7B-Instruct} to initialize the tutor model and \texttt{Llama-3.1-8B-Instruct} as the Student model, following the setup in BigMath~\citep{bigmathrl}. As a judge, \texttt{Qwen2.5-14B-Instruct} model is used. To avoid overoptimizing on the judge model used during training, in the held-out test set, a judge from another model family is used, namely, \texttt{Gemma3-27B}.  

We compare to several tutor baselines: \texttt{Qwen2.5-7B-Instruct} without any fine-tuning, \texttt{SocraticLM}~\cite{socraticlm} as a specialized open-source tutoring model and \texttt{LearnLM} as a specialized close-source tutoring model, \texttt{GPT-4o-2024-11-20} prompted to behave like a tutor, an \textit{SFT model} which uses only accepted conversations by the judges for fine-tuning, similar to~\citet{macina2023mathdialdialoguetutoringdataset}, as well as, \texttt{MDPO}~\cite{xiong2025building} which is a multi-turn extension of DPO and is trained on all pairs of chosen and rejected conversations scored by judges, similar to~\citet{sonkar-etal-2023-class,scarlatos2025dpomathdial}.

\begin{figure*}[t]
  \centering
  \begin{subfigure}{0.32\textwidth}
    \includegraphics[width=\linewidth]{\detokenize{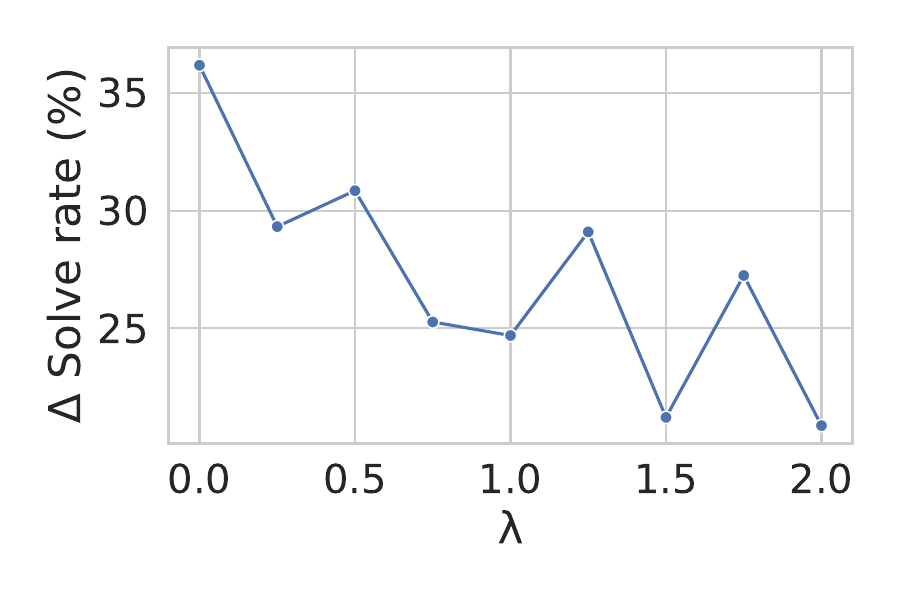}}
    \caption{$\Delta$ Solve rate vs.\ $\lambda$}
    \label{fig:delta-success}
  \end{subfigure}
  \hfill
  \begin{subfigure}{0.32\textwidth}
    \centering
    \includegraphics[width=\linewidth]{\detokenize{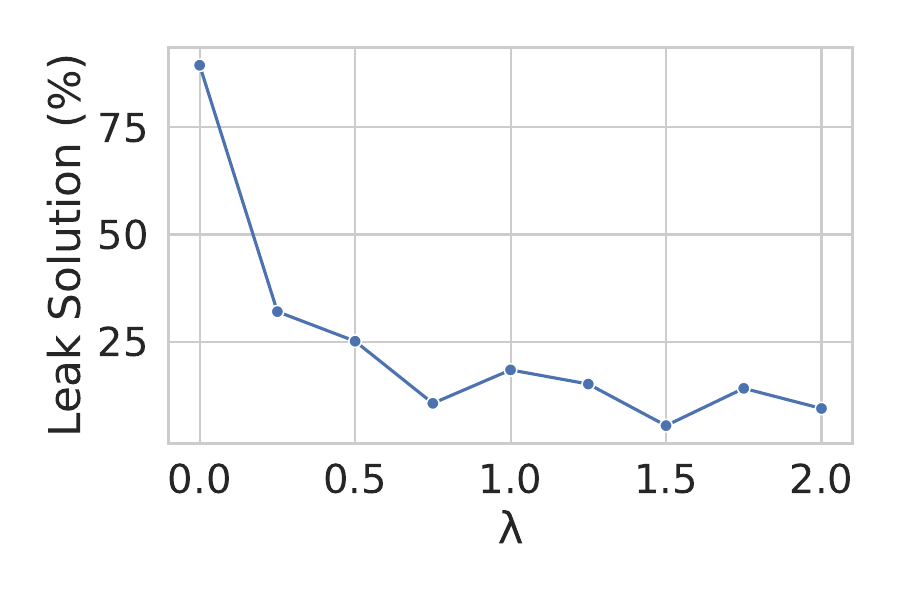}}
    \caption{Leak Solution Rate vs.\ $\lambda$}
    \label{fig:leak-sol}
  \end{subfigure}
  \hfill
  \begin{subfigure}{0.32\textwidth}
    \centering
    \includegraphics[width=\linewidth]{\detokenize{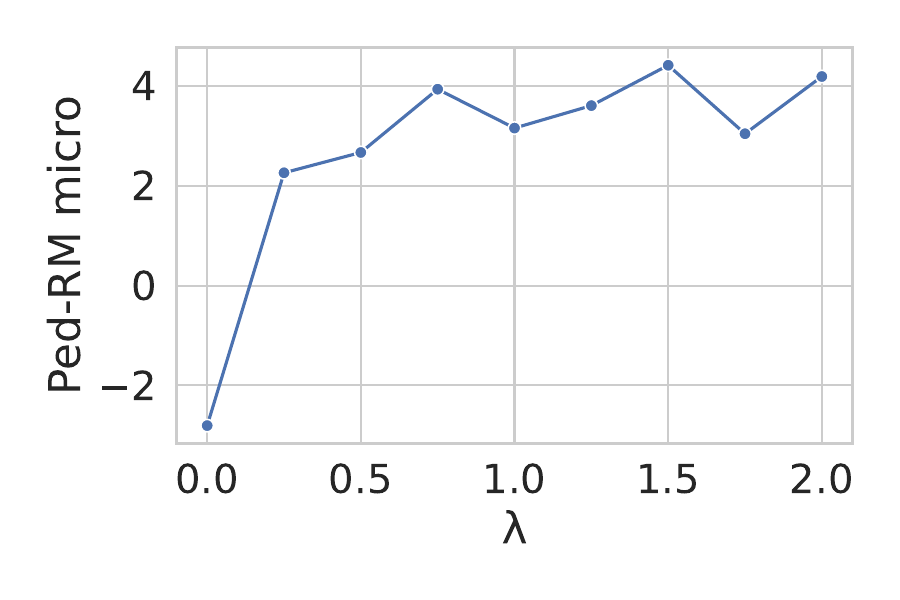}}
    \caption{Pedagogical Reward (micro) vs.\ $\lambda$}
    \label{fig:ped-rew}
  \end{subfigure}
  \caption{Performance of the RL tuned Qwen2.5-7B-Instruct across different $\lambda$ values: 
  (a) student solve rate improvement, 
  (b) leak solution rate, 
  (c) pedagogical reward (micro). 
  }
  \label{fig:all-lambda-metrics}
\end{figure*}

\begin{table*}[htbp]
    \centering
    \small
    \resizebox{\textwidth}{!}{\begin{tabular}{lccc}
        \toprule
        \textbf{Model} & \textbf{MMLU (5-shot) (\%)} & \textbf{GSM8K (4-shot) (\%)} & \textbf{MATH500 (0-shot) (\%)} \\
        \midrule
        Qwen2.5-Math-7B-Instruct              & 67.2                & \textbf{89.3}                & \textbf{81.2}                \\
        SocraticLM                            & 65.1 \textcolor{red!70!black}{(–2.1)} & 84.4 \textcolor{red!70!black}{(–4.9)} & 80.4 \textcolor{red!70!black}{(–0.8)} \\
        \midrule
        Qwen2.5-7B-Instruct                   & 77.9                & 86.8                & 75.4                \\
        Qwen2.5-7B-RL-hard--$\lambda$=1.0         & 77.3 \textcolor{red!70!black}{(-0.6)} & 86.1 \textcolor{red!70!black}{(-0.7)} & 73.6 \textcolor{red!70!black}{(-1.8)} \\
        \;\; + think         & 77.1 \textcolor{red!70!black}{(–0.8)} & 85.3 \textcolor{red!70!black}{(–1.5)} & 76.8 \textcolor{green!50!black}{(+1.4)} \\
        Qwen2.5-7B-SFT            & \textbf{79.3} \textcolor{green!50!black}{(+1.4)}                  & 79.5 \textcolor{red!70!black}{(–7.5)}                   & 66.0 \textcolor{red!70!black}{(–9.4)} \\ 
        Qwen2.5-7B-MDPO            & 78.0 \textcolor{green!50!black}{(+0.1)}                  & 87.0 \textcolor{green!50!black}{(+0.2)}                   & 76.4 \textcolor{green!50!black}{(+1.0)} \\ 
        \bottomrule
    \end{tabular}}
    \caption{Performance comparison of tutor models on MMLU, GSM8K, and MATH500 benchmarks, showing the impact of different tutor alignment strategies. SocraticLM is finetuned from Qwen2.5-Math-7B-Instruct and exhibits performance degradation relative to the original model. In contrast, our RL models finetuned from Qwen2.5-7B-Instruct demonstrate reduced degradation.
    Pedagogical-SFT, which applies supervised fine-tuning on data generated by our tutor pipeline, results in noticeable degradation. This highlights the benefits of RL-based alignment. 
    }
    \label{tab:performance_degradation}
\end{table*}

\begin{table*}[t]
  \centering
  \small
  \setlength{\tabcolsep}{4pt}
  \renewcommand{\arraystretch}{1.2}
  \resizebox{\textwidth}{!}{\begin{tabular}{l|cc|ccc|cccc}
\toprule
\multirow{3}{*}{} &
  \multicolumn{2}{c|}{\textbf{Math Expertise}} &
  \multicolumn{3}{c|}{\textbf{Student Understanding}} &
  \multicolumn{4}{c}{\textbf{Pedagogy}} \\
\cmidrule(lr){2-3}\cmidrule(lr){4-6}\cmidrule(l){7-10}
 & \multirow{2}{*}{\hdr{Problem}{solving}}
 & \multirow{2}{*}{\hdr{Socratic}{questioning}}
 & \multirow{2}{*}{\hdr{Solution}{correctness}}
 & \multirow{2}{*}{\hdr{Mistake}{location}}
 & \multirow{2}{*}{\hdr{Mistake}{correction}}
 & \multicolumn{4}{c}{Teacher response generation} \\[0.5ex]   %
 & & & & & &
   scaff.\! &
   ped.IF &
   \shortstack{scaff.\\\relax [hard]} &
   \shortstack{ped.IF\\\relax [hard]} \\
\cmidrule(lr){2-3}\cmidrule(lr){4-6}\cmidrule(l){7-10}
 & accuracy & bleu & F1 & micro F1 & accuracy &
  \multicolumn{4}{c}{win rate over human teacher} \\
\midrule

    Qwen2.5‑7B‑Instruct  & \textbf{0.87} & 0.23 & 0.63 & 0.39 & 0.04 & 0.37 & 0.60 & 0.45 & 0.56 \\
    Qwen2.5‑7B-SFT  & 0.77 & \textbf{0.24} & 0.27 & \textbf{0.45} & \textbf{0.10} & \textbf{0.64} & 0.58 & 0.57 & 0.59 \\
    Qwen2.5‑7B-MDPO  & 0.86 & 0.23 & 0.62 & 0.39 & 0.03 & 0.37 & 0.60 & 0.47 & 0.56 \\
    \midrule
    Qwen2.5‑7B-RL--$\lambda$=0.0  & 0.86 & \textbf{0.24} & 0.65 & 0.36 & 0.07 & 0.39 & 0.62 & 0.48 & 0.60 \\
    Qwen2.5‑7B-RL--$\lambda$=0.75  & 0.79 & 0.23 &  0.64 & 0.36 & 0.04 & 0.48 & 0.70 & 0.54 & 0.65 \\
    Qwen2.5‑7B-RL--$\lambda$=1.25  & 0.83 & 0.23 & \textbf{0.67} & 0.35 & 0.05 & 0.57 & \textbf{0.72} & \textbf{0.61} & \textbf{0.69} \\

    \bottomrule
  \end{tabular}}
  \caption{Results on the independent MathTutorBench benchmark with nine tasks. Scaff. and ped. IF the Scaffolding and Pedagogical Instruction Following tasks. [Hard] refers to the data split of the benchmark.
  }
  \label{tab:mathtutorbench_main}
\end{table*}

\section{Results}

\subsection{In-Domain Comparison}

\paragraph{LLMs prioritize answering over teaching}
Table~\ref{tab:result_vs_multi} presents results across model categories on an in-domain test set. Overall, we observe a trade-off between student success measured by $\Delta$ Solve rate, solution leakage and pedagogical quality, measured by Ped-RM. Qwen2.5-72B-Instruct and DeepSeek V3 achieve the highest gains in student solve rate but also exhibit high solution leakage. Qualitative example reveals that models tend to solve the problem directly for the student, see Figure~\ref{fig:example_bad_deepseek}. This supports our hypothesis that, even with engineered prompts, standard LLMs are inherently optimized for answering rather than teaching.

\paragraph{Tutoring models show improved pedagogy}
Specialized tutoring models in Table~\ref{tab:result_vs_multi}, such as, SocraticLM, SFT, and MDPO demonstrate a more balanced behavior as shown by reduced solution leakage and improved pedagogical scores. However, they often also have lower student success rates, similar to unfinetuned Qwen2.5-7B-Instruct.
The specialized, proprietary tutoring model LearnLM2.0 achieves the highest pedagogical scores while maintaining minimal leakage, indicating strong adherence to pedagogical principles. However, its low $\Delta$ solve rate suggests that it might overpenalize leaking which limits its effectiveness when students require more direct guidance.

\paragraph{Student success and pedagogy are a trade-off}
Our RL framework enables dynamic control over this trade-off. As shown in Figure~\ref{fig:all-lambda-metrics}, increasing the penalty $\lambda$ reduces solution leakage and improves pedagogical reward, at the cost of student success. Figure~\ref{fig:pareto_frontier} shows how various settings of our framework trace a Pareto frontier between student learning gains and pedagogy. At $\lambda=0.75$, for instance, our Qwen2.5-7B-RL model achieves a balanced performance across all three metrics. When $\lambda=0$, the model maximizes student success but does so by leaking answers and scoring negatively on pedagogy. 
Qualitative comparison in Figure~\ref{fig:example_good} and Figure~\ref{fig:qual_teacher_telling} further reveals that low-pedagogical-penalty models often exploit shortcuts, such as directly stating solutions or using answer fragments (e.g., `` 2+3=? ``), even if prompted not to do so. This highlights the importance of our framework when optimizing LLMs as tutors.

\paragraph{Large tutoring LLMs can be matched without human annotations}
Our online RL framing of the multi-turn dialog tutoring task trains tutoring models through interaction with a synthetic student without the need for costly human annotation. It enables scalable, multi-turn optimization with control over pedagogical behaviour via verifiable reward and LLM judge constraints. Table~\ref{tab:result_vs_multi} shows that despite using only a 7B model, our RL-tuned models (e.g. with $\lambda=1.5$ or $-r_{\text{sol}}$) outperform specialized closed-source LearnLM models on student solve rates, while nearly matching the solution leakage.

Compared to baselines using fine-tuning via SFT or preference-optimization MDPO (multi-turn extension of DPO), our approach (using $\lambda > 0$) achieves lower solution leakage and better trade-off between tutoring efficacy and student independence. This highlights the value of modeling tutoring as a multi-turn, interactive process rather than using static offline responses.

\paragraph{Thinking tags allow human observability}
Table~\ref{tab:result_vs_multi} shows that the ablation with thinking tags (+think) leads to slightly improved performance as the corresponding model without it. We observe that thinking tags allow the model to solve the problem (Figure~\ref{fig:thinking_tag_qual}) or enable the model to plan how to explain mistakes to the student (Figure~\ref{fig:thinking_tag_dice}). This is similar to what has been shown to improve model responses in previous work~\cite{stepwise}, but in our case, the model learns this behaviour during training.

\subsection{Comparison on the Out-of-Domain Data}
\paragraph{No degradation of solving capabilities}
Unlike prior approaches such as SocraticLM~\cite{socraticlm}, which sacrifice base model performance in pursuit of pedagogical alignment, our method preserves reasoning abilities across standard benchmarks. As shown in Table~\ref{tab:performance_degradation}, Qwen2.5-7B-RL matches or slightly exceeds the performance of its base model (Qwen2.5-7B-Instruct). In contrast, SocraticLM, which is fine-tuned from the Math version of Qwen, degrades performance. Similarly, supervised fine-tuning (SFT) results in decrease on math-heavy benchmarks (–7.5\% on GSM8K, –9.4\% on MATH500). These findings demonstrate that RL-based alignment better preserves core reasoning skills, avoiding the trade-off between pedagogical behaviour and task competence.

\paragraph{Out-of-domain tuturing benchmark}
Table~\ref{tab:mathtutorbench_main} shows evaluation of our models on the out-of-domain MathTutorBench benchmark~\cite{MathTutorBench}, which assesses tutoring ability on nine tasks and uses the Ped-RM to find win-rate over human teachers. Our RL-aligned 7B models match or exceed the pedagogical quality of baseline models. However, SFT remains a strong baseline for Mistake location and Mistake correction tasks, highlighting the need to carefully combine SFT and RL to build robust tutoring models in the future.

\section{Conclusion}
In this work, we propose methods to align LLMs for pedagogy using reinforcement learning.
Our method does not require human annotations beyond initial problem statements and train on the models' own context which reduces train and test mismatch.
Rewards allow balancing student solving accuracy and pedagogy, which requires strategically withholding information while accuracy could trivially be increased by the tutor leaking the solution.
We find that smaller models trained with this approach can match large, proprietary models in various tutoring metrics.

\section*{Limitations}
Our online RL approach introduces additional complexity compared to simpler SFT or single-turn pairwise preferences such as DPO. In particular, as known from other RL tasks, the use of model rollouts to simulate interactions with a student introduce variance and can make training potentially unstable or sample-inefficient. Careful implementation is required to maintain stability.

Our current reward focuses on conversation-level rewards, for example enabling to focus on longer-term post-dialog student success. However, truly learning a topic is measured with a delayed post-test on student transfer, i.e. the ability to transfer the learned topic over time. Future work could focus on such more precise but very delayed signal. 

All experiments focus on math-based tutoring tasks. While math is a valuable testbed with enough existing datasets, it represents only one STEM subject. 

Our approach trains tutoring models using interactions with a single student model only, which may not reflect the diversity of real learners. Incorporating additional student models and different student personas in a prompt could lead to more realistic settings better representing a diversity of real learners and their misconceptions.

All student responses and reward signals in our framework is generated synthetically by sampling from LLMs. While this enables scalable and cost-efficient training, it has not been validated with real students, which future works can explore, for example the impact of a trade-off between student success and pedagogy.

\section*{Ethics Statement}
\paragraph{Intended Usage} We will release the code under CC-BY-4.0 license. We use the BigMath, GSM8k, and MATH500 datasets released under the MIT license, the MathTutorBench benchmark released under CC-BY-4.0, and the MMLU with the Apache License 2.0. We use all of the datasets within their intended usage. 

\paragraph{Potential Misuse}
The overall goal of this work is to support the community in improving LLMs at tutoring capabilities and align them with good pedagogical practice based on learning sciences. However, there are potential risks related to the reward function and reward hacking. If the reward function is redefined or an inappropriate penalty is used, the model might learn a suboptimal tutoring behaviour. Similarly, if the reward function is underspecified, the risk of model hacking the reward and finding shortcuts is present. We mitigate this by including several datasets and evaluation setups. Moreover, we share the code, hyperparameters, and the setup openly. However, before deploying the model with real students we emphasize caution, adding safeguards and proper user testing.

\section*{Acknowledgements}
Jakub Macina acknowledges funding from the ETH AI Center Doctoral Fellowship, Asuera Stiftung, and the ETH Zurich Foundation. This work was supported in part by the Swiss AI Initiative under a project (ID a04) on AI for Education. This work has been funded by the LOEWE Distinguished Chair “Ubiquitous Knowledge Processing”, LOEWE initiative, Hesse, Germany (Grant Number: \text{LOEWE/4a//519/05/00.002(0002)/81}) and by the State of Hesse, Germany, as part of the project “LLMentor: Expert-AI Coteaching of ‘Introduction to Scientific Work’” (Connectom Networking and Innovation Fund). We thank Yilmazcan Ozyurt for valuable feedback and discussions.

\bibliography{example_paper}

\appendix

\section{Implementation Details}\label{app:impl_details}

\subsection{Compute Resources}
All GRPO runs were conducted using 4$\times$A100 80GB GPUs over approximately 48 hours per run. Each run covered roughly 20\% of the training data and involved around 300 policy updates. At an estimated cost of \$2 per GPU hour, each full RL training run costs approximately \$400.

\subsection{Configuration}
We adapt the standard \texttt{GRPOTrainer} from the TRL library~\cite{vonwerra2022trl} to support our multi-agent tutor-student interaction setting. For each problem instance \( P \), we randomly select one of the two supported tutoring scenarios in our environment---either \emph{student-initiated} or \emph{tutor-initiated}---and apply it uniformly across all rollouts in the corresponding batch. To compute the student solve rate, we set $K=8$. All dialog rollouts start from an empty dialog history and only problem $P$ as input.
\\
The key hyperparameters are:
\begin{itemize}[itemsep=1pt, topsep=1pt]
    \item Learning rate: $5 \times 10^{-7}$
    \item KL coefficient: $\beta = 0.001$
    \item Gradient steps per batch: $\mu = 2$
    \item Batch size: 16 problems per batch, each with 8 rollouts
    \item Sampling temperature: $\text{T}=1.0$
\end{itemize}
We use the \texttt{paged\_adamw\_8bit} optimizer \cite{8bit-optim} to reduce memory usage.

\subsection{Baselines: SFT and MDPO}
To generate data for the MDPO and SFT baselines, we sample 30\% of the full dataset and generate 8 rollouts (conversations)  $D=(\vu_1, \dots, \vu_T)$ per problem. For \textit{MDPO}, we construct within-group preference pairs \((D_{\text{acc}}, D_{\text{rej}})\) such that \(r(D_{\text{acc}}) > r(D_{\text{rej}})\), resulting in 36k preference pairs. For \textit{SFT}, we filter the MDPO data to keep only accepted responses, remove duplicates, and obtain approximately 14k accepted samples.

Training hyperparameters for baselines:
\begin{itemize}[itemsep=1pt, topsep=1pt]
    \item SFT: batch size 16, learning rate $2 \times 10^{-5}$, trained for 1 epoch
    \item MDPO: batch size 32, learning rate $2 \times 10^{-7}$, trained for 1 epoch (all settings follow the original MDPO paper~\cite{xiong2025building})
\end{itemize}

\subsection{Inference and Quantization}
To enable efficient tutor–student–judge simulation at scale, we serve all models through vLLM library~\cite{vllm}, which enables fast batched decoding with KV-caching. To reduce memory footprint and inference latency we also employ quantization. The \textit{student model} is quantized using FP8, enabling fast inference while not noticeably degrading performance. The \textit{judge model} is quantized using 4-bit Activation-Aware Quantization (AWQ)~\cite{awq}, significantly reducing compute cost.

\section{Template reward}\label{app:template-reward}
In addition to the primary pedagogical and correctness rewards, we incorporate several template-based auxiliary rewards inspired by prior work~\cite{deepseek_r1}. These rewards encourage structured and concise tutor interactions and penalize incorrect use of format tags and conversation mechanics.

\subsection{Thinking Tag Usage Reward} 

To promote transparent and interpretable internal reasoning by the tutor, we reward explicitly formatted thinking tags. Each tutor's turn can include structured reasoning enclosed within tags of the format:
\[
\texttt{<think>...</think>}
\]

We compute the reward as follows:
\[
r_{\text{think}}(\va_T\mid \vs_T) = c \times \frac{|\{\vu_i \in D \mid \vu_i \text{ correct tags}\}|}{|\{\vu_i \in D\}|},
\]
where $\vu_i$ are individual tutor utterances and c is a constant which we set to 0.5. The correct formatting implies that tags are both opened and properly closed without structural errors.

\subsection{Penalty for Incorrect Thinking Tag Formatting}
To enforce the correctness of thinking tag formatting and ensure structured output, we penalize the model for each incorrectly formatted or unclosed thinking tag:
\[
p_{\text{misuse}}(\va_T\mid \vs_T) = c \times (\text{\# of wrong tags in } D).
\]
This includes scenarios where:
\begin{itemize}
    \item A thinking tag is opened but not closed.
    \item A thinking tag is malformed or incorrectly structured.
\end{itemize}

\subsection{End-of-Conversation Reward}
To encourage the tutor model to efficiently and naturally conclude dialogs, we reward the explicit use of the special termination tag:
\[
\langle \text{end\_of\_conversation}\rangle
\]

Only the tutor is permitted to terminate the conversation by generating this special token. The reward is defined as:
\[
r_{\text{end}}(\va_T\mid \vs_T) =
\begin{cases}
0.1, & \text{if dialog is ended early}\\[3pt]
0, & \text{otherwise}.
\end{cases}
\]
This incentivizes concise, purposeful interactions, discouraging overly long dialogs.

\subsection{Penalty for Exceeding Max Tokens per Turn}

We set a maximum number of tokens allowed per tutor turn. If any tutor's turn exceeds this limit (thus failing to generate the EOS token within the maximum length), we apply a fixed penalty:
\[
p_{\text{len}}(\va_T\mid \vs_T) =
\begin{cases}
0.5, & \text{ no EOS token generated
}\\[3pt]
0, & \text{otherwise}.
\end{cases}
\]

This penalty ensures the tutor generates concise and complete responses without truncation, promoting conversational coherence.

\subsection{Combined Template Reward}

The combined auxiliary reward incorporating all these components is:
\begin{align*}
r_{\text{templ}}(\va_T\mid \vs_T) &= r_{\text{think}}(\va_T\mid \vs_T) + r_{\text{end}}(\va_T\mid \vs_T) \\
    &\quad - p_{\text{misuse}}(\va_T\mid \vs_T) \\
    &\quad - p_{\text{len}}(\va_T\mid \vs_T).
\end{align*}

\section{Prompts}
Pre-dialog solution by a student is computed using the prompt in Figure~\ref{fig:predialog_solution_prompt} and post-dialog solution by a student using the prompt in Figure~\ref{fig:postialog_solution_prompt}. Student and tutor system prompts used during a conversation are in Figure~\ref{fig:student_prompt} and Figure~\ref{fig:teacher_prompt}.
The exact prompt for judging the leakage of the solution by a teacher model is in Figure~\ref{fig:leakage-prompt} and Figure~\ref{fig:constructive-style-prompt} shows the prompt for the helpfulness of the tutor response.

\section{Example Conversations}
Examples of the conversations from our model are in Figure~\ref{fig:example_good}, Figure~\ref{fig:qual_teacher_telling}, Figure~\ref{fig:thinking_tag_qual}, Figure~\ref{fig:example_bad_deepseek}, and Figure~\ref{fig:thinking_tag_dice}.

\tcbset{
    goodexample/.style={
        colback=green!5!white,
        colframe=green!60!black,
        title={\faCheckCircle\ \textbf{Good Example: Teacher Guides Student}},
        fonttitle=\bfseries,
        arc=2mm, boxrule=0.8pt, top=1mm, bottom=1mm, left=2mm, right=2mm
    },
    goodexample_0.75/.style={
        colback=green!5!white,
        colframe=green!60!black,
        title={\faCheckCircle\ \textbf{Good Example: Teacher Guides Student ($\text{Qwen2.5-7B-RL } \lambda=0.75$)}},
        fonttitle=\bfseries,
        arc=2mm, boxrule=0.8pt, top=1mm, bottom=1mm, left=2mm, right=2mm
    },
    badexample/.style={
        colback=red!5!white,
        colframe=red!60!black,
        title={\faTimesCircle\ \textbf{Bad Example: Teacher Solves the problem fully ($\text{Qwen2.5-7B-RL } \lambda=0.0$)}},
        fonttitle=\bfseries,
        arc=2mm, boxrule=0.8pt, top=1mm, bottom=1mm, left=2mm, right=2mm
    },
    badexample_deepseek/.style={
        colback=red!5!white,
        colframe=red!60!black,
        title={\faTimesCircle\ \textbf{Bad Example: Teacher Solves the problem fully (\text{DeepSeek-V3-0324})}},
        fonttitle=\bfseries,
        arc=2mm, boxrule=0.8pt, top=1mm, bottom=1mm, left=2mm, right=2mm
    },
    thinkexample_hard/.style={
        colback=blue!3!white,
        colframe=blue!60!black,
        title={\faLightbulb\ \textbf{Guided Reasoning with Thinking ($\text{Qwen2.5-7B-RL-hard }\lambda=1.0 \text{ +think}$)}},
        fonttitle=\bfseries,
        arc=2mm, boxrule=0.8pt
    },
    thinkexample_lambda/.style={
        colback=blue!3!white,
        colframe=blue!60!black,
        title={\faLightbulb\ \textbf{Mistake analysis in thinking tags ($\text{Qwen2.5-7B-RL }\lambda=1.5 \text{ +think}$)}},
        fonttitle=\bfseries,
        arc=2mm, boxrule=0.8pt
    },
    thinkingbox/.style={
        colback=yellow!10!white,
        colframe=yellow!60!black,
        boxrule=0.5pt,
        arc=1mm,
        left=1mm, right=1mm, top=0.5mm, bottom=0.5mm
    }
}
\newcommand{\cont}{\hfill{\footnotesize\itshape[conversation continues]}\par}

\newtcolorbox{teachermsg}{
  colback=blue!5!white,
  colframe=blue!40!black,
  boxrule=0.5pt,
  arc=1mm,
  left=1mm, right=1mm, top=0.5mm, bottom=0.5mm,
  before skip=4pt, after skip=4pt,
  fonttitle=\bfseries,
  title=\faChalkboardTeacher\ Teacher
}

\newtcolorbox{studentmsg}{
  colback=gray!10!white,
  colframe=gray!40!black,
  boxrule=0.5pt,
  arc=1mm,
  left=1mm, right=1mm, top=0.5mm, bottom=0.5mm,
  before skip=4pt, after skip=4pt,
  fonttitle=\bfseries,
  title=\faUserGraduate\ Student
}

\begin{figure*}[h]
\centering
\small
\begin{tcolorbox}[goodexample_0.75]
\textbf{Problem:}  
On a 15-meter-long path, starting from the beginning, a flower pot is placed every 3 meters. How many flower pots are there in total?

\begin{teachermsg}
Hi Alex, let's think about this problem step-by-step. We have a 15-meter path and we're placing flower pots every 3 meters, starting from the very beginning. \\

Can you identify the positions where we would place the flower pots?
\end{teachermsg}

\begin{studentmsg}
So, we start from the beginning of the path. If we place a flower pot every 3 meters, then the first pot would be at 0 meters. The next pot would be at 3 meters, then 6 meters, and so on. \\

So, the positions where we would place the flower pots are 0, 3, 6, 9, and 12 meters.
\end{studentmsg}

\begin{teachermsg}
Great observation, Alex! You've correctly identified the positions of the flower pots. Now, let's make sure we count them all and consider if there's a pot at the very end of the path. \\

Can you count how many positions you've listed, including the starting point at 0 meters? And also, think about whether there's a pot right at the 15-meter mark.
\end{teachermsg}

\begin{studentmsg}
There are 5 positions I listed: 0, 3, 6, 9, and 12 meters. \\

If we have a pot every 3 meters, shouldn't there be a pot at 15 meters as well? Because 15 divided by 3 is 5, so that would be the 6th pot.
\end{studentmsg}

\begin{teachermsg}
Exactly, Alex! You're absolutely right. There should be a pot at the 15-meter mark as well. Let's list them again to make sure we haven't missed any: \\

0 meters, 3 meters, 6 meters, 9 meters, 12 meters, and 15 meters. \\

Now, how many flower pots do you think there are in total?
\end{teachermsg}

\begin{studentmsg}
So we have the pots at 0, 3, 6, 9, 12, and 15 meters. That's 6 pots in total.
\end{studentmsg}

\begin{teachermsg}
Perfect, Alex! You've counted the flower pots correctly and arrived at the right answer on your own. Well done! \\

To summarize, there are 6 flower pots in total along the 15-meter path. \\

Great job working through this problem and arriving at the solution on your own!
\end{teachermsg}
\end{tcolorbox}
\caption{Good Example: Teacher guides the student without directly giving the answer.}
\label{fig:example_good}
\end{figure*}

\begin{figure*}[h]
\centering
\small
\begin{tcolorbox}[badexample]
\textbf{Problem:} Calculate the sum of all values of $x$ that satisfy the equation
$$(x^2 - 5x + 5)^{(x^2 + 4x - 60)} = 1.$$
Consider the conditions for $a^b = 1$ and provide the exact sum of valid $x$ values.\\

\begin{teachermsg}
Hello! I'm glad you're working on this problem. Let's break it down together.\\
We have the equation: 
$$(x^2 - 5x + 5)^{x^2 + 4x - 60} = 1.$$

For $a^b = 1$, there are three main cases to consider:
\begin{itemize}
    \item $a = 1$ (for any $b$)
    \item $a = -1$ and $b$ is even
    \item $a \neq 0$ and $b = 0$
\end{itemize}

Let's apply these cases to our equation.\\

\textbf{Case 1:} $x^2 - 5x + 5 = 1$\\
Solve: $x^2 - 5x + 4 = 0 \Rightarrow (x - 1)(x - 4) = 0$\\
So, $x = 1$ or $x = 4$\\

\textbf{Case 2:} 
[... Many rows omitted]\\

\textbf{Sum of valid $x$ values:} $1 + 4 + 2 + (-10) + 6 = 3$\\
So, the sum of all valid $x$ values is $\boxed{3}$.
\end{teachermsg}
\end{tcolorbox}
\caption{Bad Example: Teacher explains too much and gives the full solution.}
\label{fig:qual_teacher_telling}
\end{figure*}

\begin{figure*}[h]
\centering
\small
\begin{tcolorbox}[thinkexample_hard]
\textbf{Problem:} Given vectors $\vec{a} = (x - 1, 2)$ and $\vec{b} = (4, y)$, if $\vec{a} \perp \vec{b}$, then the minimum value of $9^x + 3^y$ is: \\

\begin{teachermsg}
\begin{tcolorbox}[thinkingbox]
\textbf{Thinking:} Given that the vectors are perpendicular, their dot product is zero.\\
So we can set up the equation:
$$(x - 1) \cdot 4 + 2 \cdot y = 0.$$

From there, we can express one variable in terms of the other and then substitute into:
$$9^x + 3^y.$$
This gives a single-variable function we can minimize.
\end{tcolorbox}

Can you start by finding a relationship between $x$ and $y$ using the fact that the vectors are perpendicular?\\
Remember, for two vectors to be perpendicular, their dot product must be zero.
\end{teachermsg}
\cont
\end{tcolorbox}
\caption{Example with structured reasoning and no solution leak.}
\label{fig:thinking_tag_qual}
\end{figure*}

\begin{figure*}[h]
\centering
\small
\begin{tcolorbox}[badexample_deepseek]
\textbf{Problem:} \\
A student uses a calculator to find an answer but instead of pressing the $x^2$ key presses the $\sqrt{x}$ key by mistake. The student's answer was 9. What should the answer have been if the student had pressed the correct key? Express your answer as a whole number.

\begin{teachermsg}
Alright, let's break this problem down step by step.

First, the student meant to press the ($x^2$) key (which squares a number), but instead pressed the ($\sqrt{x}$) key (which takes the square root of a number). The incorrect answer they got was 9.

So, the student took the square root of some number and got 9. Let's call the original number ($x$). That means:
\[
\sqrt{x} = 9
\Rightarrow x = 81
\]

Now, the student was supposed to square this number, so:
\[
x^2 = 81^2 = 6561
\]

But wait—that seems like a very large number. Maybe I misunderstood. Let me think differently.

The student intended to square a number, say ($y$), but instead took its square root and got 9.
\[
\sqrt{y} = 9 \Rightarrow y = 81 \Rightarrow y^2 = 6561
\]

[... Many lines omitted]

Final answer: \fbox{6561}
\end{teachermsg}

\begin{studentmsg}
.. so the correct answer is 6561. I think I understand now. If the student pressed the ($\sqrt{x}$) key by mistake and got 9, it means that the original number was actually 81, and if they had pressed the correct key, they would have gotten 6561.
\end{studentmsg}

\begin{teachermsg}
Exactly! You've got it. The key steps are:

The student took the square root of some number ($x$) and got 9:
\[
\sqrt{x} = 9 \Rightarrow x = 81
\]

If they had pressed the correct key ($x^2$), they would have squared 81:
\[
81^2 = 6561
\]

Great job understanding the problem!
\end{teachermsg}
\end{tcolorbox}
\caption{Bad Example: The model solves the entire problem directly instead of prompting the student to think through the steps.}
\label{fig:example_bad_deepseek}
\end{figure*}

\begin{figure*}[h]
\centering
\small
\begin{tcolorbox}[thinkexample_lambda]
\textbf{Problem:} \\
Roll two dice, labeled A and B, one after the other. Let $a$ and $b$ represent the numbers rolled on dice A and B, respectively. If the point $M(a, b)$ falls within the region defined by the inequality $x^2 + y^2 \leq m$ (where $m$ is a constant), and this event is denoted as event $C$, find the minimum value of $m$ to make the probability of event $C$, $P(C)$, equal to 1.

\begin{studentmsg}
Here is my attempt at this problem: To solve this problem, we need to determine the region defined by the inequality $x^2 + y^2 \leq m$ and find the minimum value of $m$ that makes the probability of event $C$ equal to 1.

Since we are rolling two dice, the possible values for $a$ and $b$ are 1 to 6. The point $M(a, b)$ can be any point in a $6 \times 6$ grid.

The inequality $x^2 + y^2 \leq m$ represents a circle with radius $\sqrt{m}$ centered at the origin (0, 0).

To make the probability of event $C$ equal to 1, the circle must cover the entire $6 \times 6$ grid. The maximum distance from the origin in the $6 \times 6$ grid is $\sqrt{36} = 6$ (when $a = 6$ and $b = 6$).

So, to cover the entire grid, the radius of the circle must be at least 6. Therefore, the minimum value of $m$ is the square of the radius, which is:

\[
m = 6^2 = 36
\]

However, this is not the final answer because we need to consider the edge cases where $a = 6$ and $b = 1$ or $a = 1$ and $b = 6$. In these cases, the distance from the origin is $\sqrt{37}$, which is greater than 6.

To cover these edge cases, we need a circle with a radius of at least $\sqrt{37}$.

\[
m = (\sqrt{37})^2 = 37
\]

Note that the probability of event $C$ is not actually 1, but we've made the circle large enough to cover all possible points in the $6 \times 6$ grid.

\[
\boxed{37}
\]
\end{studentmsg}

\begin{teachermsg}
\begin{tcolorbox}[thinkingbox]
\textbf{Thinking:} While the student's reasoning is on the right track, there's a misunderstanding about the region and the probability. The student correctly identified the need for the circle to cover the $6 \times 6$ grid but made an incorrect assumption about the edge cases. I should guide them to recheck the maximum distance from the origin to any point $(a, b)$ where $a$ and $b$ are between 1 and 6.
\end{tcolorbox}

That's a great start, Riley! You're on the right track by considering the $6 \times 6$ grid and the circle that needs to cover it. However, there's a small mistake in your reasoning.\\

Let's think about the maximum distance from the origin to any point $(a, b)$ where $a$ and $b$ are the numbers rolled on the dice.\\

Can you recheck the maximum distance and why it's important for the probability of event $C$ to be 1?
\end{teachermsg}
\cont
\end{tcolorbox}
\caption{Example where the teacher analyses the mistake of the student attempt inside the thinking tags without revealing a large part of the solution.}
\label{fig:thinking_tag_dice}
\end{figure*}

\begin{figure*}[h]
    \centering
    \small
    \begin{tcolorbox}
    Please reason step by step, and put your final answer within  \textbackslash boxed\{\}. \\
    \{\{ problem \}\}
    \end{tcolorbox}
    \caption{
    Prompt for pre-dialog student solution where \textit{problem} is a placeholder for a math problem.
    \label{fig:predialog_solution_prompt}}
\end{figure*}

\begin{figure*}[h]
    \centering
    \small
    \begin{tcolorbox}
    \{\{ conversation \}\} \\
    The conversation with the teacher has ended. \\
    It is time you create a step by step complete solution including everything. \\
    It is essential you include the final answer in the \textbackslash boxed\{\} format.\\
    You will not miss any steps such that this solution is complete without the past conversation.
    \end{tcolorbox}
    \caption{
    Prompt for post-dialog student solution, where \textit{conversation} is a placeholder for tutor-student simulated conversation.
    \label{fig:postialog_solution_prompt}}
\end{figure*}

\begin{figure*}[h]
    \centering
    \small
    \begin{tcolorbox}
    You will act as a student in a conversation with a teacher in training. 
    You will need to act as much like a student as possible. If possible do not respond with overly long messages.\\
    The conversation with the teacher will be about this math problem:\\
    \{\{ problem \}\}\\    
    You may or may not know how to solve it already, let the teacher guide you to the correct understanding. 
    You will be tested at the end and scored thus it is best if you collaborate with the teacher as it has more experience in math than you.
    \end{tcolorbox}
    \caption{
    A student system prompt used in a dialog with a teacher.
    \label{fig:student_prompt}}
\end{figure*}

\begin{figure*}[h]
    \centering
    \small
    \begin{tcolorbox}
    You are tasked with being a teacher and helping a student with a math problem.
    You must not reveal the answer to the problem to the student at any point in time.\\
    Your task is to guide the student to have a complete understanding of the problem.
    Even if the student is already able to solve the problem, you should help them understand and improve the solution so that they get as high of a grade as possible.
    If possible, do not respond with overly long responses to the student.\\
    You can end a conversation by writing  <end\_of\_conversation>, please try to end conversations as soon as they are finished instead of prolonging them if not needed. But do not end them prematurely either.\\
    Here is the math problem:\\
    \{\{ problem \}\}
    \end{tcolorbox}
    \caption{
    A teacher system prompt used during a simulated conversation.
    \label{fig:teacher_prompt}}
\end{figure*}

\begin{figure*}[h]
    \centering
    \small
    \begin{tcolorbox}
    You are tasked with inspecting a conversation between a student and a teacher (tutor). Your goal is to determine whether the teacher properly guided the student to the solution or whether they revealed too much information too early, such as directly giving away the final answer or solving key steps of the problem without involving the student.\\
    \\
    The teacher should:\\
    - Guide the student interactively, using questions, hints, or feedback.\\
    - Encourage the student to reason through the problem themselves.\\
    - Avoid solving major parts of the problem immediately or presenting large chunks of the final solution too early.\\
    - Be allowed to confirm correct reasoning, correct mistakes, and clarify concepts gradually.\\
    - Clarify certain parts in more detail if the student does not get it.\\
    \\
    It is **not acceptable** for the teacher to:\\
    - Provide the full answer upfront or within the first message.\\
    - Reveal all key steps in a single message without student involvement.\\
    - Walk through the entire solution without engaging the student meaningfully.\\
    \\
    \#\# **Examples of correct and incorrect guidance**:\\
    \#\#\# Example 1 – Good Guidance (Integration by parts):\\
    - Student: How do I integrate $x * e^{(2x)}$?\\
    - Teacher: Good question! Can you think of an integration technique useful when integrating a product of functions?\\
    - Student: Integration by parts?\\
    - Teacher: Exactly. Can you identify what you'd set as 'u' and what as 'dv' here?\\
    - Student: u = x and dv = $e^{(2x)}$ dx?\\
    - Teacher: Nice job. Can you find du and v from here?\\
    \{
      "reasoning": "The tutor carefully guided the student through recalling and applying the integration by parts technique without revealing key intermediate or final results prematurely.",
      "decision": "OK"
    \}
    \\\\
    \lbrack 1 more example reject example \rbrack\\
    \\
    \#\#\# Example 3 – Immediate Final Answer (Limit computation):\\
    - Student: How can I compute the limit as x → 0 of sin(5x)/x?\\
    - Teacher: The limit is simply \\boxed\{5\}.\\
    \\
    \{
      "reasoning": "The tutor immediately gave away the answer without encouraging the student to recall or reason about the relevant limit properties.",
      "decision": "REJECT"
    \}
    \\\\
    \lbrack 1 more ok example \rbrack\\
    \\
    \#\# **Evaluate the following conversation**:\\
    ==BEGIN CONVERSATION==\\
    \lbrack conversation \rbrack\\
    ==END CONVERSATION==\\
    \\
    Please output a JSON evaluation following exactly this schema:\\
    \{
      "reasoning": "your explanation here", "decision": "OK or REJECT"
    \}\\
    \\
    Now write the JSON evaluation for the conversation above, make sure the JSON is valid:    
    \end{tcolorbox}
    \caption{
    Prompt for judging whether the tutor leaked the answer.
    \label{fig:leakage-prompt}}
\end{figure*}

\begin{figure*}[h]
    \centering
    \small
    \begin{tcolorbox}
    Your task is to inspect a conversation between a **student** and a **teacher**.

    Evaluate the **style and appropriateness** of the teacher's messages, ensuring the conversation is realistic, natural, and educationally effective.
    \\
    \\
    \#\#\# **Acceptable Teacher Style**:\\
    - Messages are concise and easy to understand.\\
    - Teacher patiently and respectfully engages the student.\\
    - **Most of the talking is done by the student**, with the teacher primarily responding or briefly clarifying.\\
    - Conversation feels natural and believable (e.g., conversational tone, realistic student-teacher interaction).\\

    \#\#\# **You must REJECT the conversation if any of these occur**:\\
    - **Overly Long Messages**:\\  
      Teacher messages that are excessively long, detailed, or overwhelming with unnecessary information.\\
      *(Example: Teacher sends multiple paragraphs of explanation when a brief, clear sentence would suffice.)*\\
    - **Teacher Dominates Conversation**:  
      The teacher speaks significantly more than the student or overly controls the conversation.  
      *(Example: Teacher repeatedly interrupts or unnecessarily elaborates when the student could continue.)*\\
    - **Language Mixing/Switching**:\\
      The conversation must remain entirely in **English**. Switching or mixing languages—even briefly—is not allowed.\\  
      *(Example: Teacher says something partially or completely in Spanish.)*\\
    - **Unrealistic/Unnatural Interaction**:\\  
      Conversation doesn't feel believable between a real teacher and student (awkward phrasing, unnatural dialog, abrupt changes in tone).  
      *(Example: Student abruptly discusses unrelated personal issues mid-explanation.)*\\
    - **Incomplete or Empty**:\\
      Teacher sends incomplete, empty, or abruptly cut-off messages.\\  
      *(Example: Teacher's message suddenly ends mid-sentence.)*\\
    \\
    \#\#**Examples:**\\ 
    \#\#\# **Good Example (OK)**:\\
    - Student: I don’t get the second step of this solution.\\
    - Teacher: Could you tell me exactly which part confuses you?\\
    - Student: It's the part where you simplify the fraction.\\
    - Teacher: Sure! Let's focus on that simplification. Do you remember how we reduce fractions?\\
    \\
    \#\#\# **Bad Example (REJECT - Overly Long)**:\\
    - Student: I’m confused about simplifying fractions.\\
    - Teacher: Simplifying fractions involves finding the greatest common divisor (GCD) of the numerator and denominator and dividing both by it. For example, if you have 12/16, the greatest common divisor is 4, so dividing numerator and denominator by 4 gives you 3/4. Another example is 10/20; the GCD here is 10, so this reduces to 1/2. You must always look carefully to identify the largest divisor...\\
    *(Reason: Excessively long and detailed explanation.)*\\
    \\
    \#\#\# **Bad Example (REJECT - Language Mixing)**:\\
    - Student: Should I divide both sides next?\\
    - Teacher: Yes, exactly! Divide ambos lados por dos.\\
    *(Reason: Mixing English and Spanish.)*\\
    \\
    Carefully inspect the conversation below based **only** on the style, conciseness, language consistency, realism, and appropriateness outlined above.\\
    **Conversation to evaluate**:\\
    ==BEGIN CONVERSATION==\\
    \lbrack conversation \rbrack\\
    ==END CONVERSATION==\\
    \\
    Please output a JSON evaluation following exactly this schema:\\
    \{
      "reasoning": "your explanation here", "decision": "OK or REJECT"
    \}\\
    \\
    Now write the JSON evaluation for the conversation above, make sure the JSON is valid:    
    \end{tcolorbox}
    \caption{
    Prompt for judging helpfulness which consists of constructive support and teacher tone.
    \label{fig:constructive-style-prompt}}
\end{figure*}

\end{document}